\documentclass[letterpaper, 10 pt, conference]{ieeeconf}  

\IEEEoverridecommandlockouts                              

\usepackage{graphicx}
\usepackage[table]{xcolor}
\usepackage{tcolorbox}
\usepackage{hhline}
\usepackage{hyperref}
\usepackage[margin=0.8in]{geometry}
\usepackage{caption}
\usepackage{subcaption}

\title{\LARGE Fast Object Segmentation Learning with Kernel-based Methods for Robotics 
}

\author{Federico Ceola$^{1,2}$, Elisa Maiettini$^{1}$, Giulia Pasquale$^{1}$, Lorenzo Rosasco$^{2,3}$ and Lorenzo Natale$^{1}$
\thanks{$^{1}$Federico Ceola, Elisa Maiettini, Giulia Pasquale and Lorenzo Natale are with Humanoid Sensing and Perception, Istituto Italiano di Tecnologia, Genoa, Italy
        {\tt\footnotesize name.surname@iit.it}}
\thanks{$^{2}$Lorenzo Rosasco and Federico Ceola are with Laboratory for Computational and Statistical Learning and with Dipartimento di Informatica, Bioingegneria, Robotica e Ingegneria dei Sistemi, University of Genoa, Genoa, Italy}
\thanks{$^{3}$Lorenzo Rosasco is also with Istituto Italiano di Tecnologia and Massachusetts Institute of Technology, Cambridge, MA {\tt\footnotesize lrosasco@mit.edu}}
}

\begin{document}
\begin{minipage}{\textwidth}
  \copyright~2021 IEEE. Personal use of this material is permitted. Permission from IEEE must be obtained for all other uses, in any current or future media, including reprinting/republishing this material for advertising or promotional purposes, creating new collective works, for resale or redistribution to servers or lists, or reuse of any copyrighted component of this work in other works.
  \end{minipage}

\maketitle
\thispagestyle{empty}
\pagestyle{empty}


\begin{abstract}
Object segmentation is a key component in the visual system of a robot that performs tasks like grasping and object manipulation, especially in presence of occlusions. Like many other computer vision tasks, the adoption of deep architectures has made available algorithms that perform this task with remarkable performance. However, adoption of such algorithms in robotics is hampered by the fact that training requires large amount of computing time and it cannot be performed on-line.\\
In this work, we propose a novel architecture for object segmentation, that overcomes this problem and provides comparable performance in a fraction of the time required by the state-of-the-art methods. Our approach is based on a pre-trained Mask R-CNN, in which various layers have been replaced with a set of classifiers and regressors that are re-trained for a new task. We employ an efficient Kernel-based method that allows for fast training on large scale problems.\\
Our approach is validated on the YCB-Video dataset which is widely adopted in the computer vision and robotics community, demonstrating that we can achieve and even surpass performance of the state-of-the-art, with a significant reduction ($\mathbf{{\sim}6\times}$) of the training time.\\
The code to reproduce the experiments is publicly available on GitHub\footnote{\url{https://github.com/robotology/online-detection}}.

\end{abstract}




\section{INTRODUCTION}
\label{sec:introduction}
Object segmentation is a key task in computer vision and robotics.
A precise localization of the object on the image plane is key to solve 
problems like pose estimation \cite{xiang2018posecnn, Zakharov_2019_ICCV, Wang_2019_CVPR}, refinement \cite{Li_2018_ECCV, Manhardt_2018_ECCV} and grasping or manipulation \cite{deng2020}.\\
Like many other computer vision tasks (e.g., image classification and object detection), for decades this was mostly addressed with feature extraction (e.g., sparse coding \cite{yang2009}, Fisher vectors \cite{perronin2010}) followed by Kernel methods \cite{vedaldi2009} or structured prediction approaches \cite{krahenbuhl2011efficient, lucchi2012}. In recent years, instead, end-to-end training of deep architectures became mainstream since it showed to provide superior performance \cite{Long2015, He2017, minaee2020}.\\
However, deep learning is known to require large training data to optimize complex architectures with expensive and long training procedures.
While the problem of training data has recently been alleviated by synthetic image generation \cite{Remez_2018_ECCV, blenderproc}, the computational burden remains.\\
Lately, the literature of one-shot and few-shot object segmentation in video \cite{Caelles_arXiv_2019} presented methods based on the idea of fine-tuning only part of the network on-line \cite{DAVIS20175th, DAVIS20178th}. This approach can learn to segment novel objects in shorter time than standard end-to-end approaches, while keeping performance. However, the procedure of fine-tuning a deep network in general does not guarantee generalization properties, since it involves a non convex optimization and several hyper-parameters, while it can be still prohibitive for robotic settings which require learning to happen very fast (e.g., in seconds or minutes).\\
Recently, it was shown that pre-trained deep features can be efficiently re-used also to train Kernel-based methods for tasks like image classification \cite{Chatfield2014_lvgg} object detection and even segmentation \cite{farabet, girshick2014_rcnn}. 
The advantage of Kernel-based methods, among others, is that the number of hyper-parameters to tune is smaller (e.g. in~\cite{falkon2018}, one regularization parameter, one Kernel parameter) and convergence and generalization properties are guaranteed \cite{falkon2018}. 
This approach was recently adopted to achieve high performance and remarkably faster training times in interactive object learning applications for object recognition \cite{pasquale2019} and detection \cite{maiettini2019a} in robotics. By relying on the above pipelines, a robot can observe a short image sequence of the objects of interest and seamlessly train to detect them right afterwards\footnote{\label{video_oor}\url{https://youtu.be/HdmDYIL48H4}}\textsuperscript{,}\footnote{\label{video_ood}\url{https://youtu.be/eT-2v6-xoSs}}.\\
In this paper, we build on this prior work and present a method to re-use general-purpose deep features (e.g., pre-trained on COCO \cite{coco}) to train Kernel methods for the \textit{dense} task of image segmentation. 
As evidenced from prior work \cite{maiettini2019a, girshick2014_rcnn, farabet}, the main challenge lies in the difficulty of training a \textit{non-linear} classifier over the \textit{large} training sets that characterize object detection and segmentation problems (where each sample is a region or a single pixel in an image).\\
To this end, we feed deep features to FALKON~\cite{falkon2018}, a recently proposed Nystr{\"o}m-based Kernel method optimized for large-scale problems. We instantiate one per-pixel FALKON classifier for every class and train it on pixel-wise features extracted from the second-last layer of the network.\\
We first solve the detection task by applying the same approach, which was presented in \cite{maiettini2019a}. Then, we feed the dense features extracted from the detected regions to FALKON classifiers to perform figure-ground segmentation within each region. 
To tackle the well-known problem of dataset (positive-negative) imbalance which characterizes the task of object detection, we use the bootstrapping technique from \cite{maiettini2019a}. Instead, to manage the great quantity of data in the figure-ground segmentation task, we apply data subsampling in our training procedure.\\
As network, we use the region-based Mask R-CNN \cite{He2017}, to which we basically replace every output layer with a set of FALKON classifiers or regressors.
The resulting architecture is clean and flexible, allowing to easily train accurate instance segmentation on small datasets in a few minutes, but possibly also on large ones.\\
We benchmark our method on the YCB-Video dataset \cite{xiang2018posecnn}, a state-of-the-art benchmark for pose estimation and tracking in challenging conditions for robotics, featuring clutter and occlusion across varied objects.
In fact, while pose estimation results are widespread over this latter, but also other robotics datasets (see, e.g., the BOP challenge \cite{hodan2020bop}), it is less common for robotic papers to present the partial results over the segmentation task --which, however, is almost always part of their pipelines.
Hence, we also hope that this work could pave the way to other works in robotics, to present not only pose estimation but also intermediate segmentation results.\\
The remaining of this paper is organized as follows. In Sec.~\ref{sec:relwork} we review the state-of-the-art in the fields of instance segmentation and efficient learning strategies for visual tasks in robotics. Then in Sec.~\ref{sec:methods}, we describe the proposed approach and in Sec.~\ref{sec:experiments} we report on the performed empirical validation. Finally, in Sec.~\ref{sec:conclusions}, we summarize the main contributions and identify directions for future works.

\section{RELATED WORK}
\label{sec:relwork}
In this section, we overview the main state-of-the-art approaches for object segmentation (Sec.~\ref{sec:relwork:segmentation}), then we present the literature that tackles the problem of fast adaptation of robotic vision systems to novel tasks (Sec.~\ref{sec:relwork:ood}). 

\subsection{Object instance segmentation}
\label{sec:relwork:segmentation}
The task of \textit{instance segmentation}~\cite{minaee2020} aims at classifying all the pixels in an image, while identifying different instances of the categories of interest. Note that, this task is different from \textit{semantic segmentation}, which has the similar objective of classifying every pixel of an image but not being aware of the different instances. While the field of semantic segmentation is mainly dominated by approaches based on Fully Convolutional Networks (FCNs)~\cite{Long2015,Shelhamer2017}, the literature in instance segmentation can be divided in the following three groups. 

\noindent{{\bf Region-based approaches.}} Methods in this group split the segmentation task into four main stages: (i) a convolutional feature map is generated by a FCN applied to the whole image, (ii) a list of Regions of Interest (RoIs) is generated by a region proposal method, (iii) a RoI pooling layer~\cite{ren2015_faster} warps each RoI and extracts per-RoI features from the convolutional map and, finally, (iv) two sibling shallow Neural Networks predict detections and masks for each feature map. The main representative example of this group is Mask R-CNN~\cite{He2017} which extends the object detection architecture Faster R-CNN~\cite{ren2015_faster} by adding, in parallel to the final detection layers, a further output layer for binary mask prediction. Note that, this is in contrast to latest approaches~\cite{Dai2016b,Li2017,Pinheiro2015}, where the classification depends on the prediction of the masks. Mask R-CNN is devised as a monolithic architecture which trains end-to-end, via stochastic gradient descent and back-propagation, all the components for region proposals and feature extraction, object detection and segmentation. The network optimization is carried out by minimizing a ``multi-task loss"~\cite{ren2015_faster,He2017}, which accounts for the contribution of classification, bounding box and mask precision (see~\cite{ren2015_faster,He2017} for details). 

\noindent{{\bf Pixel-based approaches.}} Methods of this group predict a so-called ``auxiliary" information for each pixel (like, e.g, energy levels~\cite{bai2017}) and then a clustering algorithm groups pixels into object instances based on such information. An example of this approach is presented in~\cite{bai2017}, where the watershed transform is combined with a deep convolutional network to produce an energy map of the image where object instances
are represented as energy basins. Then, they perform a cut at a single energy level to yield connected components corresponding to object instances. 

\noindent{{\bf Snake approaches.}} Methods of this group deform an initial coarse contour to the object boundary by optimizing a handcrafted or learned energy measure with respect to the contour coordinates. In~\cite{ling2019}, for instance, the presented pipeline uses a graph convolutional network to predict vertex-wise offsets for contour deformation. In~\cite{peng2020}, instead, the object's contour is not treated as a general graph, but it leverages the cycle graph topology and uses the circular convolution for efficient feature learning on a contour.\\

\noindent{All the aforementioned methods of the state-of-art are ``monolithic'' deep architectures, trained end-to-end via back-propagation and stochastic gradient descent, with long training times that are not suited for on-line robotic applications. In this work, we propose a region-based method for object segmentation that, building on previous work~\cite{maiettini2019a}, addresses this issue, allowing for fast model learning.}

\begin{figure*}
	\centering
	\includegraphics[width=0.8\linewidth]{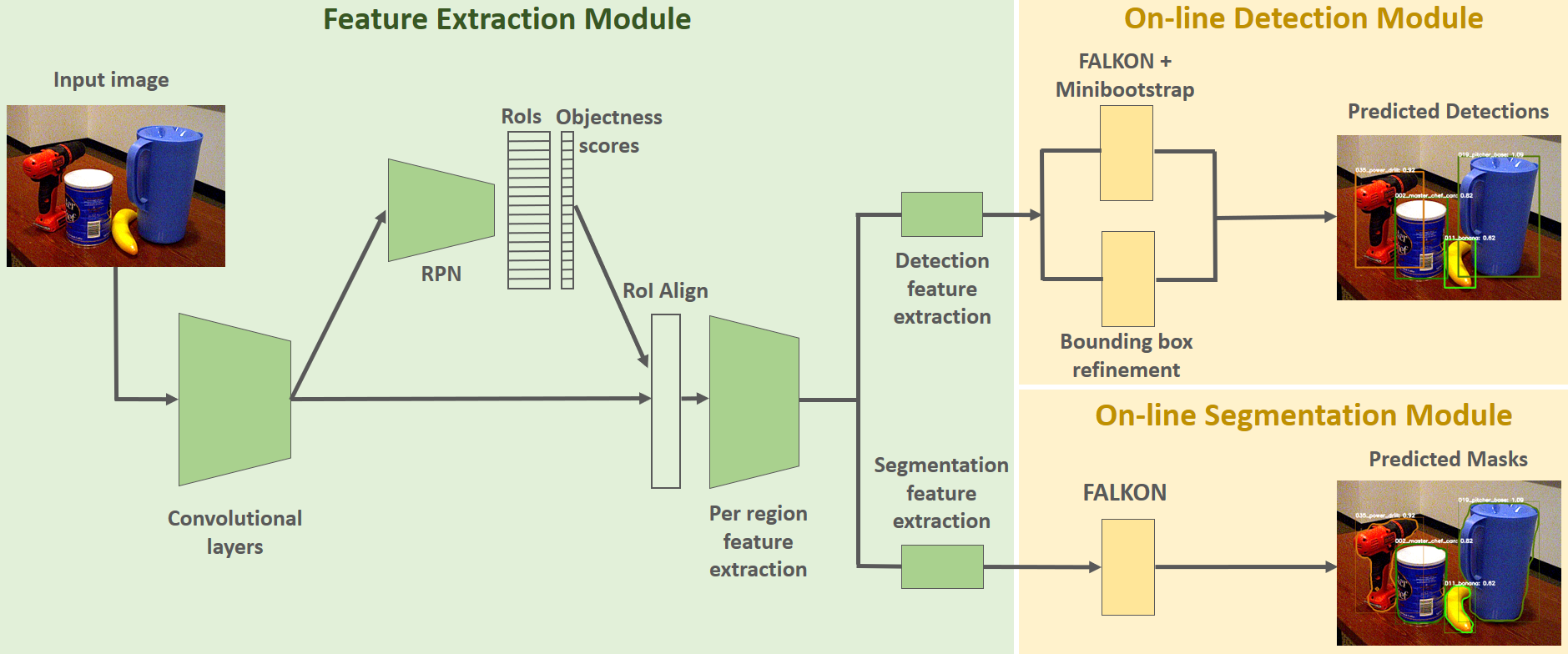}
	\caption{Overview of the proposed pipeline. The \textit{Feature Extraction Module} is composed of Mask R-CNN's first layers trained off-line on the FEATURE-TASK. For each input image, it extracts per RoI features to train on-line on the TARGET-TASK the \textit{On-line Detection Module} and the \textit{On-line Segmentation Module}. The \textit{On-line Detection Module} classifies the RoIs and predicts the corresponding detections. The \textit{On-line Segmentation Module} predicts masks for each detection. 
	Note that the picture represents the pipeline at training time. At test time, the detection output is fed as input to the RoI Align.}
	\label{fig:pipeline}
\end{figure*}

\subsection{Fast object detection methods in robotics}
\label{sec:relwork:ood}
Fast and efficient adaptation to novel tasks is critical for robots that need to operate in unconstrained environments. Previous works in robotics analyze this problem and propose solutions for object recognition~\cite{pasquale2019} and object detection~\cite{maiettini2018,maiettini2019a}. In both cases, the proposed pipelines rely on hybrid architectures that integrate deep Convolutional Neural Network's (CNN) feature descriptors with efficient ``shallow" Kernel-based methods (like, e.g., FALKON~\cite{falkon2018}) for respectively classification or detection. In these works, the key aspect to achieve on-line training time is to decouple the learning of the features and region proposals extractor from the optimization of the Kernel-based methods. Moreover, in~\cite{maiettini2019a} an approximated bootstrapping procedure for selecting hard negative samples has been proposed, to address the well-known foreground-background imbalance problem in object detection.\\
In this work, we extend the pipeline proposed in~\cite{maiettini2019a}, to also perform object segmentation. Following the same principle as in~\cite{maiettini2019a}, in our proposed approach, the training of the features and region proposals extractor is decoupled by the optimization of the detection and segmentation methods, allowing for a fast model re-adaptation.

\section{METHODS}
\label{sec:methods}
In this work, we present an instance segmentation method, which allows to learn to predict masks of previously unseen objects (referred to as TARGET-TASK) in a fraction of the time required by state-of-the-art approaches. To do this, it relies on some components of a deep learning based instance segmentation  network trained once and off-line on the available data, depicting a different set of objects (referred to as FEATURE-TASK).\\
In this section, we describe the proposed approach. Sec.~\ref{methods:overview} provides an overview of the pipeline, while in Sec.~\ref{methods:onlinelearning} the proposed on-line learning strategy for instance segmentation is described.

\subsection{Overview of the pipeline}
\label{methods:overview}
The pipeline proposed in this work is mainly composed of three modules, as depicted in Fig.~\ref{fig:pipeline}:
\begin{itemize}
  \item The \textbf{Feature Extraction Module} employs the first layers of the Mask R-CNN architecture, trained on the FEATURE-TASK, to extract two sets of convolutional features for a set of RoIs, one for \textbf{On-line Detection} and one for \textbf{On-line Segmentation}.
  \item The \textbf{On-line Detection Module} uses the per-region features computed by the feature extractor to predict their class and to refine their bounding box with the approach described in~\cite{maiettini2019a}.
  \item The \textbf{On-line Segmentation Module}, instead, predicts the masks of the instances contained in the RoIs with the approach proposed in this work and described in Sec.~\ref{methods:onlinelearning}. As the previous module, also this module is trained on the TARGET-TASK.
\end{itemize}


\noindent{The} main contribution relies on the latter module. Indeed, we propose a fast and efficient learning strategy that allows to update the instance segmentation model in few minutes. Moreover, we integrate it in the on-line detection pipeline presented in~\cite{maiettini2019a}, extending it and proposing a novel approach. This permits fast adaptation time as new data is available for both object detection and segmentation.

\noindent{{\bf Learning protocol.}}
Our learning procedure is composed of two stages. Firstly, Mask R-CNN is trained off-line on the FEATURE-TASK via backpropagation, following the training protocol proposed in~\cite{He2017}. Then, the on-line training of both the detector and the mask predictor on the TARGET-TASK are performed in three sub-steps. First, some of the layers of the Mask R-CNN architecture learned in the first phase on the FEATURE-TASK (specifically the backbone, the RPN and the convolutional layers following the Roi Align, depicted as green blocks in Fig.~\ref{fig:pipeline}), are used to jointly extract the two sets of features needed to train the \textit{On-line Detection Module} and the \textit{On-line Segmentation Module} (represented by yellow blocks in Fig.~\ref{fig:pipeline}). These two modules are then sequentially trained with the on-line learning strategy described in  Sec.~\ref{methods:onlinelearning}.

\subsection{On-line learning strategy}
\label{methods:onlinelearning}

As mentioned in Sec.~\ref{methods:overview}, the trainings of the \textit{On-line Detection Module} and of the \textit{On-line Segmentation Module} are performed by employing two different sets of features computed by the \textit{Feature Extraction Module}. These are extracted in two different points of the Mask R-CNN architecture. Specifically, features necessary to train the \textit{On-line Detection Module} are extracted from the set of one-dimensional tensors, which in Mask R-CNN are given as input to the last fully connected layers of the detection head. Features necessary to train the \textit{On-line Segmentation Module}, instead, are contained in the convolutional feature maps produced as output by the penultimate convolutional layer of the mask head in Mask R-CNN. Once the sets of features are extracted for all the training images, the \textit{On-line Detection Module} and the \textit{On-line Segmentation Module} can be trained with the procedures described below.

\noindent{{\bf On-line Detection.}}
For the training of the \textit{On-line Detection Module}, we adopt the method described in~\cite{maiettini2019a}, but we use the improved implementation of the \textit{Feature Extraction Module} proposed in~\cite{ceola2020fast}. Indeed, such improvement allows to increase the quality of the features, due to the substitution of the first layers of Faster R-CNN~\cite{ren2015_faster}, with the ones from Mask R-CNN~\cite{He2017}. As in~\cite{maiettini2019a}, the last fully connected layers of the detection head in Mask R-CNN are substituted with $N$ FALKON~\cite{falkon2018,falkonlibrary2020} binary classifiers ($N$ being the number of classes of the TARGET-TASK) for class prediction, and by $N\times4$ Regularized Least Squares (RLS) regressors for the refinement \cite{girshick2014_rcnn} of the RoIs proposed by the RPN. To deal with the well-known problem of foreground-background imbalance and with the high redundancy among the negative examples in object detection problems~\cite{Lin2017focal}, the $N$ FALKON classifiers are trained with the \textit{Minibootstrap} proposed in~\cite{maiettini2019a}. This procedure is an adaptation of the Hard Negatives Mining strategy 
\cite{girshick2014_rcnn} and it allows to train the classifiers in a few seconds by visiting a random subset of negative examples.

\noindent{{\bf On-line Segmentation.}}
In Mask R-CNN~\cite{He2017}, the mask head is a shallow Fully Convolutional Network (FCN), which for every RoI computes $N$ squared masks (one for each of the $N$ classes) of fixed size $s \times s$. In particular, the last layer of the segmentation branch (i.e., the per-pixel mask predictor) is a convolutional layer with $N$ channels, kernel size $1$ and stride $1$ that receives as input a convolutional feature map of size $s\times s \times f$ (where $f$ is the number of channels). Therefore, the output tensor of the segmentation head has dimension $s \times s \times N$ and the $ijn^{th}$ element of such tensor represents the probability of the pixel in the location $ij$ of belonging to an instance of the $n^{th}$ class.
In this work, we substitute the $N$ kernels for per-pixel classification with $N$ FALKON binary classifiers. We flatten the $s \times s \times f$ feature map into a list of $s\times s$ feature tensors of size $f$ associated to specific locations in the RoIs and use them as samples to train the FALKON classifiers. At training time, we consider the ground-truth bounding boxes as RoIs. The $ij^{th}$ element of the input feature map of size $s\times s \times f$ is considered as a positive training sample for the $n^{th}$ classifier if the $ij^{th}$ pixel of the associated ground-truth mask belongs to the foreground and the ground-truth class is $n$. Otherwise, if it is a background pixel within a ground-truth bounding box of class $n$, it is considered as a negative sample for the $n^{th}$ classifier. To speed-up the training procedure, both the positive and the negative instances are subsampled by a fixed factor $r$. We employ FALKON to address the per-pixel classification problem since, as shown in~\cite{maiettini2019a} (specifically, refer to Tab. 6 in~\cite{maiettini2019a}), among other equivalent choices, it allows to achieve the best time-accuracy trade-off (refer to~\cite{falkon2018} for details).

\noindent{{\bf Hyper-parameters.}}
For both the detection and segmentation modules, we use FALKON with a Gaussian Kernel.
The main hyper-parameters of the proposed method are:
\begin{itemize}
    \item The standard deviation $\sigma$ of the Gaussian Kernel, the regularization parameter $\lambda$  and the number of Nystr{\"o}m centers $M$ of the FALKON classifiers of both the \textit{On-line Detection Module} and the \textit{On-line Segmentation Module}.
    \item The \textit{Minibootstrap} parameters to train the \textit{On-line Detection Module}, i.e., the number of batches $n_B$ and the size of such batches $BS$.
    \item The sampling factor $r$ of the training examples in the \textit{On-line Segmentation Module}.
\end{itemize}
For our experiments, we rely on the criterion proposed in~\cite{maiettini2019a} to set the \textit{Minibootstrap}'s parameters, we cross-validate the FALKON's parameters as explained in Sec~\ref{experiments:setup} and we report on an empirical analysis of the impact of the sampling factor $r$.

\section{EXPERIMENTS}
\label{sec:experiments}
In this section, we report the experiments performed to validate the proposed approach. In Sec.~\ref{experiments:setup}, we describe the experimental setup. In Sec.~\ref{experiments:benchmark}, we provide the results obtained on the YCB-Video benchmark. Finally, in Sec.~\ref{experiments:ablation}, we report on ablation studies to analyze our approach. 

\subsection{Experimental setup}
\label{experiments:setup}

\begin{table*}[]
	\centering
	\begin{tabular}{c|c|c|c|c|c}
		\cline{1-6}
		\centering \textbf{Method}                                             & \textbf{mAP50 bbox(\%)}    & \textbf{mAP50 segm(\%)}   & \textbf{mAP70 bbox(\%)}    & \textbf{mAP70 segm(\%)}   & \textbf{Train Time} \\ \hline
		\multicolumn{1}{c|}{Mask R-CNN~\cite{He2017} (output layers)}  		   &        \textbf{76.87}      & 74.12                     &68.56              & 64.80                     & 1h 57m 1s           \\ \hline
		\rowcolor[HTML]{A8E4A0} 
		\multicolumn{1}{c|}{Ours}                      &        76.46        & \textbf{74.66}              &\textbf{68.71}                & \textbf{64.96}              & \textbf{19m 32s}        \\ \hline
	\end{tabular}
	\caption{Benchmark on the \textsc{YCB-Video} dataset.  We compare the accuracy achieved by (\textbf{Mask R-CNN (output layers)}) (first row) and by the proposed approach (\textbf{Ours}) (second row).}
	\label{table:benchmark}
\end{table*}

\begin{figure*}
	\centering
	\includegraphics[width=0.9\linewidth]{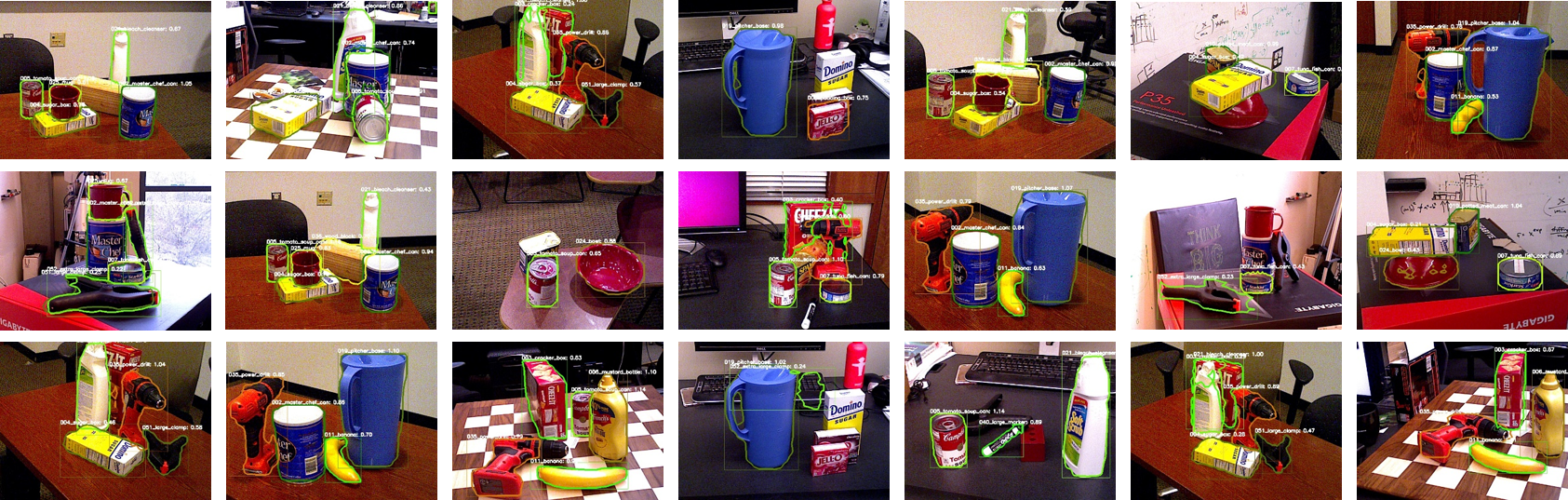}
	\caption{Randomly chosen predictions of \textbf{Ours} on the YCB-Video $keyframe$ set.}
	\label{fig:qualitative}
\end{figure*}

In this work, we consider Mask R-CNN as baseline for comparison with the proposed pipeline (\textbf{Ours}). We start with Mask R-CNN trained on the FEATURE-TASK and we fine-tune the output layers on the TARGET-TASK (i.e., the fully connected layers for detection and the last convolutional layer for masks prediction). We refer to this architecture and training method as \textbf{Mask R-CNN (output layers)}.
We use ResNet50~\cite{He2015} as backbone of Mask R-CNN for both the \textbf{Mask R-CNN (output layers)} baseline and the \textit{Feature Extraction Module} of the proposed approach (\textbf{Ours}). To set the training hyper-parameters, for both methods we perform a cross-validation on a validation set. Specifically, the model chosen for the \textbf{Mask R-CNN (output layers)} baseline has been trained for the minimum number of epochs at which the highest segmentation accuracy is achieved on the validation set. In \textbf{Ours}, instead, the validation set is used to select $\sigma$s and $\lambda$s of the FALKON classifiers, both in the \textit{On-line Detection Module} and in the \textit{On-line Segmentation Module}. To compare the performance between the two approaches, we consider three metrics: the mean Average Precision (mAP), as defined in~\cite{pascal2010} of the predicted detections (\textbf{mAP bbox(\%)}) and of the predicted masks (\textbf{mAP segm(\%)}) and the training time. Concerning the bounding boxes and masks IoU to be considered as positives matches in the mAP computations, we set the thresholds to $50\%$ and $70\%$. As regards the training time\footnote{ All the experiments reported in this paper have been performed on a machine equipped with Intel(R) Xeon(R) E5-2690 v4 CPUs @2.60GHz, and a single NVIDIA(R) Tesla P100 GPU.}, for \textbf{Mask R-CNN (output layers)} it is the time necessary to fine-tune Mask R-CNN via backpropagation. For \textbf{Ours}, instead, it is composed of the feature extraction time and of the training time of the on-line modules. The former one in practical robotic applications can happen during the data acquisition phase\textsuperscript{\ref{video_oor},\ref{video_ood}}, which can be performed, for instance, as in~\cite{10.3389/frobt.2016.00035, maiettini2017, xie2020best, suchi2019easylabel}, where the ground-truth is collected with automatic procedure. 


\noindent{{\bf Datasets.}}
In our experiments we consider the YCB-Video~\cite{xiang2018posecnn} dataset. Specifically, we use as training set a sample of the 80 training sequences, by selecting one every ten frames, resulting into a set of 11320 images. For testing, instead, we use the 2949 images included in the \textit{keyframe} set. To select the training hyper-parameters, according to the training protocol described in~Sec.\ref{experiments:setup}, we use a set of 1000 images, randomly sampled from the 12 test sequences, which are not included in the \textit{keyframe} set. The 80 categories in the MS COCO~\cite{coco} dataset, instead, compose the FEATURE-TASK in some of our experiments.

\subsection{Benchmark on the YCB-Video dataset}
\label{experiments:benchmark}

We validate the proposed on-line segmentation approach by considering the 21 objects from the YCB-Video dataset as TARGET-TASK and MS COCO as FEATURE-TASK. As described in Sec.~\ref{experiments:setup}, we compare the performance obtained by our approach (\textbf{Ours}), with the \textbf{Mask R-CNN (output layers)} baseline. As in~\cite{maiettini2019a}, to train the \textit{On-line Detection Module} we set the \textit{Minibootstrap}'s batch size $BS$ to $2000$, but, given the higher amount of training images with respect to the benchmark reported in~\cite{maiettini2019a} (Table 2), we increase the number of batches $n_B$ to $15$. The number of Nystr{\"o}m centers $M$ of the FALKON classifiers of both the \textit{On-line Detection Module} and the \textit{On-line Segmentation Module} is set to $2000$. The sampling factor $r$ of the segmentation training samples is set to $0.3$ (refer to Sec.~\ref{experiments:ablation} for an analysis of the impact of this parameter).\\
Results reported in Tab.~\ref{table:benchmark} show that our approach, while being $\sim$6 times faster than the baseline, outperforms \textbf{Mask R-CNN (output layers)} in the segmentation accuracy computed with both IoU thresholds, while achieving comparable performance in terms of detection mAP. Train time of \textbf{Ours} is composed of $18$min:$23$s for feature extraction, $35$s to train the \textit{On-line Detection Module} and $34$s to train the \textit{On-line Segmentation Module}. This shows that, once the features are extracted, the TARGET-TASK can be learned extremely fast. We report in Fig.~\ref{fig:qualitative}, some random images from the test set with the overlaid predicted masks.

\subsection{Ablation studies}
\label{experiments:ablation}
 
In this section, we provide three ablation studies of the proposed pipeline to further validate our approach. Firstly, we evaluate segmentation accuracy by decoupling such task from the detection one. Then, we analyze the impact on accuracy and training time of the sampling factor $r$ (see Sec.~\ref{methods:onlinelearning}). Finally, to compare the accuracy of our method with the one of Mask R-CNN fully trained on the TARGET-TASK, we benchmark our approach in a scenario in which FEATURE-TASK and TARGET-TASK correspond.

\begin{figure*}
     \centering
     \begin{subfigure}[b]{0.33\textwidth}
         \centering
         \includegraphics[width=\textwidth]{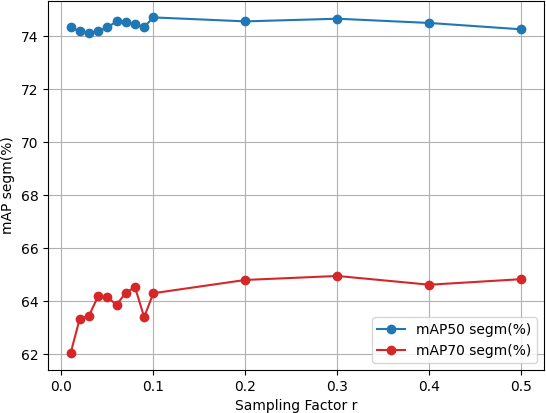}
         \caption{$ $}
         \label{fig:sf_vs_map}
     \end{subfigure}
     \begin{subfigure}[b]{0.33\textwidth}
         \centering
         \includegraphics[width=\textwidth]{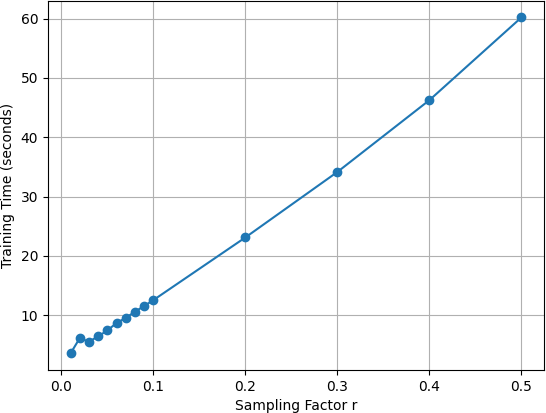}
         \caption{$ $}
         \label{fig:sf_vs_tt}
     \end{subfigure}
     \caption{Segmentation mAPs and training time for increasing values of the sampling factor $r$.}
     \label{fig:abl_sampling_factor}
\end{figure*}

\begin{table*}[]
	\centering
	\begin{tabular}{c|c|c|c|c}
		\cline{1-5}
		\centering \textbf{Method}                                             & \textbf{mAP50 bbox(\%)}    & \textbf{mAP50 segm(\%)}   & \textbf{mAP70 bbox(\%)}   & \textbf{mAP70 segm(\%)}   \\ \hline
		\multicolumn{1}{c|}{Mask R-CNN~\cite{He2017} (full)}  		           &        89.76               & 91.29                     &86.05                      & \textbf{80.41}                   \\ \hline
		\rowcolor[HTML]{A8E4A0}
		\multicolumn{1}{c|}{Ours}                      &        \textbf{90.98}      & \textbf{92.28}            &\textbf{86.90}             & 80.17             \\ \hline
	\end{tabular}
	\caption{Ablation study with \textsc{YCB-Video} dataset used as both FEATURE-TASK and TARGET-TASK. We compare the results of the full training of Mask R-CNN (\textbf{Mask R-CNN (full)}) (first row) with the proposed approach (\textbf{Ours}) (second row) considering as feature extractor the weights learned with the training in the first row.}
	\label{table:ycbv-ycbv}
\end{table*}

\noindent{{\bf Segmentation of ground-truth bounding boxes.}} In Mask R-CNN~\cite{He2017} and in our pipeline, the segmentation performance depends on the detection. To separately evaluate the segmentation errors, we consider the ideal case in which the bounding boxes proposed by the detection branch correspond to the ground-truths. FEATURE-TASK and TARGET-TASK are the same as the ones considered in Sec.~\ref{experiments:benchmark}. The achieved segmentation mAP is \textbf{94.66} and \textbf{81.78} when the IoU thresholds are set respectively to $50\%$ and $70\%$. We compare the obtained results with \textbf{Mask R-CNN (last layers)} as in Sec.~\ref{experiments:benchmark} where, as for \textbf{Ours}, we consider the ground-truth boxes as proposed detections. For the baseline, the segmentation mAP is \textbf{93.70} and \textbf{79.46} when the IoU thresholds are set to $50\%$ and $70\%$, showing the effectiveness of the proposed \textit{On-line Segmentation Module}.

\noindent{{\bf Training samples vs. training time and mAP.}} Since the time required to train the \textit{On-line Segmentation Module} is critical in robotic applications, we provide an analysis of how the choice of the sampling factor $r$ affects both the training time and the segmentation accuracy of such module. As it can be observed in Fig.~\ref{fig:sf_vs_map}, the sampling factor $r$ can be pushed to the extreme while preserving the segmentation accuracy in terms of both the considered IoU thresholds. By using a small fraction of training examples, the training times can be reduced until they become in the order of a few seconds, as it can be noticed in Fig.~\ref{fig:sf_vs_tt}. When the sampling factor $r$ is $0.01$, the number of per class positives and negatives training samples is ranging from ${\sim}1k$ to ${\sim}4k$.

\noindent{{\bf YCB-Video for both FEATURE-TASK and TARGET-TASK.}} The goal of this work is to quickly learn to segment previously unseen objects. To this aim, in Sec.~\ref{experiments:benchmark} we showed the efficiency of the proposed approach when the TARGET-TASK is different form the FEATURE-TASK. Nevertheless, to further validate our pipeline, we evaluate the performance of our method when the two tasks correspond. In particular, we set them by considering the 21 object identification task in the YCB-Video dataset (the TARGET-TASK defined in Sec.~\ref{experiments:setup}). In this case, we train, as baseline, the entire Mask R-CNN network (starting from the weights pre-trained on the COCO dataset) on the TARGET-TASK (\textbf{Mask R-CNN (full)}). Then, we use such weights to extract the features needed to train the \textit{On-line Detection Module} and the \textit{On-line Segmentation Module} with the approach described in Sec.~\ref{methods:onlinelearning} (\textbf{Ours}). The obtained accuracy are reported in Tab.~\ref{table:ycbv-ycbv}. Since the full training of Mask R-CNN on the TARGET-TASK is necessary both to obtain a baseline for our comparison and subsequently to train our pipeline, for this experiment the training times cannot be compared. Results reported in Tab.~\ref{table:ycbv-ycbv} show that, as in previous experiments, our approach provides comparable performance with respect to the baseline in terms of detection and segmentation mAP with both the IoU thresholds.

\section{CONCLUSIONS}
\label{sec:conclusions}
Fast learning and efficient adaptation are fundamental for robots that need to quickly adjust their vision systems to ever-changing environments.
In this perspective, this work presents a system for instance segmentation that extends and improves previous on-line learning approaches for object detection~\cite{maiettini2019a}. The proposed method combines Mask R-CNN for feature extraction and two sets of FALKON classifiers to efficiently perform object detection and segmentation. The resulting pipeline allows for fast adaptation to novel tasks in a fraction of the time of the deep learning based counter parts, representing a step toward the implementation of more adaptive robotic vision systems.\\
In Sec.~\ref{experiments:ablation}, we showed that we can achieve high per-pixel prediction accuracy with few training samples. We believe that this can be pushed to the extreme by integrating few-shots learning techniques in the proposed pipeline. Moreover, given the importance of instance segmentation for other more complex tasks, such as 6D pose estimation and object grasping, the proposed approach may represent a starting point towards the on-line adaptation of such tasks.



\bibliographystyle{unsrt}
\bibliography{bibliography,bibliography_elisa}  

\end{document}